\documentclass[11pt,a4paper]{article}
\usepackage{acl2013}
\usepackage{times}
\usepackage{latexsym}
\usepackage{url}
\usepackage{color}
\usepackage{multirow}
\usepackage{comment}
\usepackage{amsmath} 
\usepackage{amssymb}

\usepackage{etoolbox}
\usepackage{graphicx}
 
 \usepackage{enumitem}
  {\begin{itemize}[topsep=0pt, partopsep=0pt] %
    \setlength{\itemsep}{0pt}%
    \setlength{\parskip}{0pt}%
    }%
  {\end{itemize}}
  
  \usepackage{color}
  {\begin{enumerate}≈ \footnotesize%
    \setlength{\itemsep}{0pt}%
    \setlength{\parskip}{0pt}%
    }%
  {\end{enumerate}}
  
  \usepackage{CJKutf8}

  \newenvironment{tenumerate}%
  {\begin{enumerate}[topsep=0pt, partopsep=0pt] %
    \setlength{\itemsep}{0pt}%
    \setlength{\parskip}{0pt}%
    }%
  {\end{enumerate}}
  
\usepackage{soul}
\title{Reflections on Sentiment/Opinion Analysis}

\author{Jiwei Li$^1$ and Eduard Hovy$^2$\\
{\normalsize$^1$Computer Science Department, Stanford University, Stanford, CA 94305, USA}\\
{\normalsize$^2$Language Technology Institute, Carnegie Mellon University, Pittsburgh, PA 15213, USA}\\
{\normalsize jiweil@stanford.edu~~~~~~hovy@cmu.edu}
}

\begin{document}
\maketitle

\section{Introduction}
\label{sec:intro}

Sentiment analysis is an application of natural language processing that focuses on identifying expressions that reflect authors' 
opinion-based attitude (i.e., good or bad, like or dislike) toward entities (e.g., products, topics, issues) or facets of them (e.g., price, quality). 

Since the early 2000s, a large number of models and frameworks have been introduced to address this application, 
with emphasis on various aspects like opinion related entity exaction, review mining, topic mining, sentiment summarization, recommendation, 
and these extracted from significantly diverse text sources including product reviews, news articles, social media (blogs, Twitter, forum discussions), and so on. 

However, despite this activity, disappointingly little has been published about what exactly a {\em sentiment} or {\em opinion} actually is.  
It is generally simply assumed that two (or perhaps three) polar values {\em positive, negative, neutral}) are enough, 
and that they are clear, and that anyone would agree on how to assign such labels to arbitrary texts.  
Further, existing methods, despite employing increasingly sophisticated (and of course more powerful) models (e.g., neural nets), 
still essentially boil down to considering individual or local combinations of words and matching them against predefined lists of words with fixed sentiment values, 
and thus hardly transcend what was described in the early work by Pang et al.\ \shortcite{pang2002thumbs}.  

There is nothing against simple methods when they work, but they do not always work.
The goal of this paper is to identify why sometimes they do not work, and where to go next, 
We try to identify gaps in the current sentiment analysis literature and to outline practical computational ways to address these issues.

\paragraph{Goals, Expectations and Sentiments.}
We begin with the fundamental question ``What make people hold positive attitudes towards some entities and negative attitudes toward others?". 
The answer to this question is a psychological state 
that relates to the opinion holder's satisfaction and dissatisfaction with some aspect of the topic in question.  
One of only two principal factors determines the answer: 
either (1) the holder's deep emotionally-driven, non-logical {\bf native preferences}, 
or (2) whether (and how well) one of the holder's {\bf goals} is fulfilled, and how (in what ways) the goal is fulfilled. 

Examples of the former are reflected in sentences like ``I just like red" or ``seeing that makes me happy".  
They are typified by adverbs like ``just" and ``simply" that suggest that no further conscious psychological reflection or motivation obtains.  
Of this class of factor we can say nothing computationally, and do not address it in the rest of this chapter.  

Fortunately, a large proportion of the attitudes people write about reflect the other factor, which one can summarize as goal-driven utility.  
This relates primarily to Consequentialism: 
both to Utilitarianism, in which pleasure, economic well-being and the lack of suffering are considered desirable, 
but also to the general case that morally justifiable actions (and the objects that enable them) are desirable.  
That is, the ultimate basis for any judgment about the rightness or wrongness of one's actions, and hence of the objects that support/enable them, is a consideration of their outcome, or consequence.  

In everyday life, people establish and maintain goals or expectations, both long-term or short-term, urgent or not-urgent, ones.  
Achieving these goals would fill one with satisfaction, otherwise dissatisfaction: 
a man walks into a restaurant to achieve the goal of getting full, he cannot be satisfied if all food was sold out (the main goal not being achieved).
A voter would not be  satisfied if his candidate or party fails to win an election, since the longer-term consequences would generally work against his own preferences.
The generation of sentiment-related texts is guided by such sorts of mental satisfaction and dissatisfaction induced by goals being achieved or needs being fulfilled.  

We next provide some examples to illustrate why identifying these aspects is essential and fundamental for adequate sentiment/opinion analysis.  
Following the most popular motivation for computational sentiment analysis, suppose we wish to analyze customers' opinions towards a product or an offering.  
It is not sufficient to simply determine that someone likes or dislikes something; 
to make that knowledge useful and actionable, one also wants to know {\em why} that is the case.  
Especially when one would like to change the opinion, it is important to determine what it is about the topic that needs to be changed.  

\paragraph{Case (1)}
\begin{itemize}
\item {\bf Question}: {\it Why did the customer like detergent X?}
\item {\bf Customer's review}: {\it The detergent removes stubborn stains.}
\end{itemize}
No general sentiment indicator is found in the above review.  But the review directly provides the reason, and assuming his/her goal of clean clothing is achieved, it is evident that the opinion holder holds a positive opinion towards the detergent. 
\paragraph{Case (2)}
\begin{itemize}
\item {\bf Question}: {\it Why did the traveller dislike flight Y?}  
\item {\bf Customer's review}: {\it The food was good.  The crew was helpful and took care of everything.  The service was efficient.  However the flight was supposed to to take 1.5 hours but was 3 hours late, and I missed my next connecting flight.} 
\end{itemize}
The major goal of taking a flight is to get to your destination, which is more important than goals like enjoying one's food and receiving pampering service.  While multiple simultaneous goals induce competing opinion decisions, the presence of an importance ranking among them determines the overall sentiment.   
\paragraph{Case (3)}
\begin{itemize}
\item {\bf Question}: {\it Why did the customer visit restaurant Z?} 
\item {\it {\bf Review1}: The food is bad.}
\item {\it {\bf Review2}: The waiter was kind but the food was bad.}
\item {\it {\bf Review3}: The food was good but the waiter was rude.}
\end{itemize}
Although the primary goal of being sated may be achieved, secondary goals such as enjoying the food and receiving respectful service can be violated in various combinations.  Often, these goals pertain to the method by which the primary goal was achieved; in other words, to the question ``how?" rather than ``why?".  

A sentiment determination algorithm that can provide more than just a simple opinion label thus has to pay attention both to the primary reason behind the holder's involvement with the topic (``why?") and to the secondary reasons (both ``why?" and ``how?"), and has to be able to determine their relative importance and relationship to the primary goal.  

\paragraph{Goals and Expectations are Personal.}
As different people (opinion holders) are from different backgrounds, have different personalities, and are in different situations, 
they have different goals, needs, and the expectations of life.  
This diversity generally leads to completely diverse opinions towards the same entity, the same action, and the same situation: 
a billionaire wouldn't be the least bit concerned with the price in a bread shop but would consider the quality, 
while a beggar might care only about the price. 
This rather banal observation is explained best by Maslow's famous hierarchy of needs \cite{maslow1943theory}, 
in which the beggar's attention focuses on Maslow's Physiological needs while the billionaire's focuses on Self-Actualization; more on this in Section~\ref{sec:maslow}.  

\paragraph{Life Requires Trade-offs.} 
Most situations in real life address many personal needs simultaneously. 
People thus face trade-offs between their goals, which entails sacrificing the achievement of one goal for the satisfaction of another.
Given the variability among people, the rankings and decision procedures will also from individual to individual.  
However, Maslow's hierarchy describes the general behavioral trends of people in most societies and situations.  

\paragraph{Complex Sentiment Expressions.} 
As far as we see, current opinion analysis frameworks mostly fail to address the kinds of issues mentioned above, 
and thereby impair a deeper understanding about opinion or sentiment.  
As a result, they find it impossible to provide even rudimentary approaches to cases such as the following (from \cite{hovy2015sentiment}): 
\begin{enumerate}
\item {\it Peter thinks the pants are great and I cannot agree more.} 
\item {\it Peter thinks the pants are great but I don't agree}. 
\item {\it Sometime I like it but sometimes I hate it}. 
\item {\it He was half excited, half terrified}.
\item {\it The movie is indeed wonderful, but for some reason, I just don't like it}. 
\item {\it Why I won't buy this game even though I like it. } 
\end{enumerate}
In this paper, we explore the feasibility of addressing these issues in a practical way using machine learning techniques currently available.

\section{A Review of Current Sentiment Analysis}
\label{sec:review}

Here we give a brief overview of tasks in current sentiment analysis literature. More details can be found in \cite{liu2010sentiment,liu2012sentiment}. 

The key points involved at the algorithm level in the sentiment analysis literature follow the basic approaches of statistical machine learning, 
in which a gold-standard labeling of training data is obtained through manual annotation or other data harvesting approaches (e.g., semi-supervised or weakly supervised), and this is then used to train a variety of association-learning techniques who are then tested on new material.  
Usually, some text unit has to be identified and then associated with a sentiment label (e.g., positive, neutral, negative).  
Based on the annotated dataset, the techniques learn that vocabulary items like ``bad", ``awful", and ``disgusting" are negative sentiment indicators while ``good", ``fantastic" and ``awesome" are positive ones.  
The main complexity lies in learning which words carry some opinion and, especially, what to decide in cases where different words with opposite labels appear in the same clause.  

Basic sentiment analysis identifies the simple polarity of a text unit (e.g., a token, a phrase, a sentence, or a document) and is framed as a binary or multi-class classification task; see for example Pang et al 's work \shortcite{pang2002thumbs} that uses a unigram/bigram feature-based SVM classifier. 
Over the past 15 years, techniques have evolved from simple rule-based word matching to more sophisticated 
feature and signal (e.g., local word composition, facets of topics, opinion holder) identification and combination, 
from the level of single tokens to entire documents, 
and from `flat' word strings without any syntactic structure at all to incorporation of complex linguistic structures (e.g., discourse or mixed-affect sentences); 
see  \cite{pang2004sentimental,hu2004mining,wiebe2005annotating,nakagawa2010dependency,maas2011learning,tangbuilding,qiu2011opinion,wang2012baselines,tangjoint,yang2014context,snyder2007multiple}.  
Recent progress in neural models provides new techniques for local composition of both opinion and structure (e.g., subordination, conjunction) 
using distributed representations of text units (e.g., \cite{socher2013recursive,irsoy2014deep,irsoy2014opinion,tang2015sentiment,tang2014learning}). 

A supporting line of research extends the basic sentiment classification to include related aspects and facets, 
such as identifying opinion holders, the topics of opinions, topics not explicitly mentioned in the text, etc.; see \cite{choi2006joint,kim2006extracting,kim2004determining,li2014sentiment,jin2009novel,breck2007identifying,johansson2010syntactic,yang2012extracting,yang2013joint,yang2014joint}.  
These approaches usually employ sequence labeling models (e.g., CRF \cite{lafferty2001conditional}, HMM \cite{liu2004text}) to identify whether the current token corresponds to a specific sentiment-related aspect or facet. 

An important part of such supportive work is the identification of the relevant aspects or facets of the topic (e.g., the ambience of a restaurant vs.\ its food or staff or cleanliness) and the correspondent sentiment; see \cite{brody2010unsupervised,lu2011multi,titov2008joint,jo2011aspect,xueke2013aspect,kim2013hierarchical,garcia2013retrieving,wang2011latent,moghaddam2012design}.
Online reviews (about products or offerings) in crowdsourcing and traditional sites (e.g., yelp, Amazon, Consumer Reports) include some sort of aspect-oriented star rating systems where more stars indicate higher level of satisfaction.  Consumers  rely on these user-generated online reviews when making purchase decisions.  
To tackle this issue, researchers invent aspect identification or target extraction approaches as one subfield of sentiment analysis.  These approaches first identify 'aspects/facets of the principal Topic and then discover authors' corresponding opinions for each one; e.g., \cite{brody2010unsupervised,titov2008joint}. 
Aspects are usually identified either manually or automatically using word clustering models (e.g., LDA \cite{blei2003latent} or pLSA).  
However, real life is usually a lot more complex and much harder to break into a series of facets (e.g., quality of living, marriage, career). 

Other related work includes opinion summarization, aiming to summary sentiment key points given long texts
(e.g., \cite{hu2004mining,liu2005opinion,zhuang2006movie,ku2006opinion}), 
opinion spam detection aiming at identifying fictitious reviews generated to deceive readers
(e.g., \cite{ott2011finding,litowards,li2013identifying,jindal2008opinion,lim2010detecting}), 
sentiment text generation 
(e.g., \cite{mohammad2011once,blair2008building}), 
and large-scale sentiment/mood analysis on social media for trend detecion 
(e.g., \cite{o2010tweets,bollen2011twitter,conover2011political,paul2011you}).

\section{The Needs and Goals behind Sentiments}

As outlined in Section~\ref{sec:intro}, this chapter argues that an adequate and complete account of utilitarian-based sentiment is possible only with reference to the goals of the opinion holder.  In this section we discuss a classic model of human needs and associated goals and then outline a method for determining such goals from text.  

\subsection{Maslow's Hierarchy of Needs}
\label{sec:maslow}

\begin{figure} 
\centering
\includegraphics[width=3in]{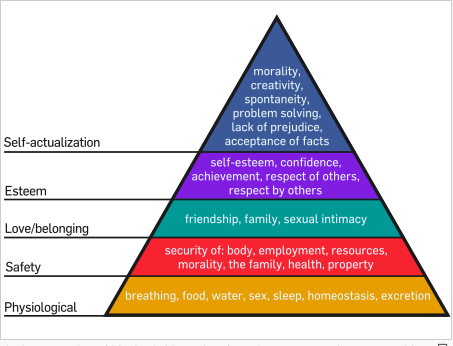}
\caption{Maslow's Hierarchy of Needs. Figure borrowed from Wikipedia \url{https://en.wikipedia.org/wiki/Abraham_Maslow}}\label{fig1}
\end{figure}

Abraham Maslow \cite{maslow1943theory,maslow1967theory,maslow1970motivation,maslow1972farther} developed a theory of the basic human needs as being organized in a hierarchy of importance, visualized using a pyramid (shown in Figure~\ref{fig1}), where needs at the bottom are the most pressing, basic, and fundamental to human life (that is, the human will tend to choose to satisfy them first before progressing to needs higher up). 

According to Maslow's theory, the most basic two levels of human needs are\footnote{References from \\\url{https://en.wikipedia.org/wiki/Abraham_Maslow};\\ \url{https://en.wikipedia.org/wiki/Maslow's_hierarchy_of_needs};\\ \url{http://www.edpsycinteractive.org/topics/conation/maslow.html}}:
\begin{itemize}
\item Physiological needs: breathing, food, water, sleep, sex, excretion, etc.
\item Safety Needs: security of body, employment, property, heath, etc.
\end{itemize}
which are essential for the physical survival of a person. Once these needs are satisfied, people tend to accomplish more and move to higher levels:
\begin{itemize}
\item Love and Belonging: psychological needs like friendship, family, sexual intimacy. 
\item Esteem: the need to be competent and recognized such as through status and level of success like achievement, respect by others, etc. 
\end{itemize}
These four types of needs are also referred to as \textsc{deficit needs} (or \textsc{D-needs}), meaning that for any human, if he or she doesn't have enough of any of them, he or she will experience the desire to obtain them. 
Less pressing than the D-needs are the so-called \textsc{growth needs}, including Cognitive, Aesthetic (need for harmony, order and beauty), and Self-actualization (described by Maslow as ``the desire to accomplish everything that one can, to become the most that one can be").  
Growth needs are more generalized, obscure, and computationally challenging.  
We focus in this chapter on deficit needs.  
For further reading, refer to Maslow's original papers \shortcite{maslow1943theory,maslow1967theory} or relevant Wikipedia pages. 

We note that real life offers many situations in which an action does not easily align with a need listed in the hierarchy (for example, the goal of British troops to arrest an Irish Republican Army leader or of US troops to attack Iraq).  Additionally, a single action (e.g., going to college, looking for a job) can simultaneously address multiple needs. 
Putting aside such complex situations in this chapter, we focus on more tractable situations to illustrate the key points\footnote{However, putting them aside them doesn't mean that we don't need to explore and explain these complex situations.  On the contrary, these situations are essential and fundamental to the understanding of opinion and sentiment, but requires deeper and more systematic exploration in psychology, cognitive science, and AI.}.  

\subsection{Finding Appropriate Goals for Actions and Entities}
\label{sec:finding-goals}

Typically, each deficit need gives rise to one or more goals that impel the agent (the opinion holder) to appropriate action.  Following standard AI and Cognitive Science practice, we assume that the agent instantiates one or more plans to achieve his or her goals, where a plan is a sequence of actions intended to alter the state of the world from some situation (typically, the agent's initial state) to a situation in which the goal has been achieved and the need satisfied.  In each plan, its actions, their preconditions, and the entities used in performing them (the plan's so-called {\it props}) constitute the material upon which sentiment analysis operates.  
For example, the goal to {\em sate one's hunger} may be achieved by plans such as {\em visit-restaurant, cook-and-eat-meal-at-home, buy-or-steal-ready-made-food, cadge-meal-invitation}, etc.  In all these plans, {\em food} is one of the props.  For the restaurant and buying-food plans, an affordable {\em price} is an important precondition.  

A sentiment detection system that seeks to understand why the holder holds a specific opinion valence has to determine the specific actions, preconditions, and props that are relevant to the holder's goal, and to what degree they suffice.  In principle, a complete account requires the system to infer from the given text: 
\begin{enumerate}
\item what need is active, 
\item which goal(s) have been activated to address the need, 
\item which plan(s) is/are being followed to achieve the goal(s), 
\item which actions, preconditions, and props appear in these plan(s), 
\item which of these is/are being talked about in the text, 
\item how well it/they actually have furthered the agent's plan(s), 
\end{enumerate}
from which the sentiment valence can be automatically deduced.  When the valence is given in the text, one can work `backwards' to infer step 6, and possibly even earlier steps.  

Determining all this is a tall order for computational systems.  Fortunately, it is possible to circumvent much of this reasoning in practice.  For most common situations, a relatively small set of goals and plans obtains, and the relevant actions, preconditions, and props are usually quite standard.  (In fact, they are precisely what is typically called `facets' in the sentiment analysis literature, for which, as described in Section~\ref{sec:review}, various techniques have been investigated, albeit without a clear understanding of the reason these facets are important.)  

Given this, the principal unaddressed computational problem today is the determination from the text of the original need or goal being experienced by the holder, since that is what ties together all the other (and currently investigated) aspects.  How can one, for a given topic, determine the goals an agent would typically have for it, suggest likely plans, and potentially pinpoint specific actions, preconditions, and props?  

One approach is to perform automated goal and plan harvesting, using typical text mining / pattern-matching approaches from Information Extraction.  This is a relatively mature application of NLP \cite{hearst1992,riloff1997,riloff1999,snow-etal2005,davidov-rappoport2006,etzioni-etal2005,banko2009,mitchell-etal2009,ritter-etal2009,kozareva-hovy2013}, and the harvesting power and behavior of various styles of patterns has been investigated for over two decades.  (In practice, the Double-Anchored Pattern (DAP) method \cite{kozareva-hovy2013} works better than most others.)  Stated simply, one creates or automatically induces text patterns anchored on the topic (e.g., a camera) such as 
\begin{quotation} 
  \noindent 
  ``I want a camera because * "\\
  ``If I had a camera I could * " \\
  ``the main reason to get a camera is * " \\
  ``wanted to *, so he bought a camera" \\
   etc.  
\end{quotation}
and then extracts from large amounts of text the matched VPs and NPs as being relevant to the topic.  Appropriately rephrased and categorized, one obtains the information harvested by these patterns would provide typical goals (reasons) for buying and using cameras.

\section{Toward a Practical Computational Approach}

We are now ready to describe the overall approach necessary for a more complete sentiment analysis system.  
For illustrative purposes we focus on simple binary (positive/negative) valence identification.  However, the framework applies to finer granularity (e.g., multi-class classification, regression) with minor adjustments.  We first provide an overall algorithm sketch, provide a series of examples, and then suggest models for determining the still unexplored aspects required for deeper sentiment analysis.   

First, we assume that standard techniques are employed to find the following from some given text: 
\begin{enumerate}
\item Opinion Holder: Individual or organization holding the opinion.
\item Entity/Aspect/Theme/Facet: topic or aspect about which the opinion is held.
\item Sentiment Indicator: Sentiment-related text (tokens, phrases, sentences, etc.)\ that indicate the polarity of the holder.
\item Valence: {\it like, neutral}, or {\it dislike}. 
\end{enumerate}
These have been defined (or at least used with implicit definition) throughout the sentiment literature, and are defined for example in \cite{hovy2015sentiment}.  
Of these, item 1 is usually achieved by simple matching. 
Item 2 can be partially addressed by recent topic/facet mining models, and item 3 can be addressed by existing sentiment related algorithms at the word-, sentence-, or text-level.  Item 4 at its simplest is a matter of keyword matching, but the composition witin a sentence of contrasting valences has generated some interesting researech.  
Annotated corpora (or other semi-supervised data harvesting techniques) might be needed for goal and need identification, as discussed above. 

Given this, the following sketch algorithm implements deeper sentiment analysis: \\
\rule{7.5cm}{0.03cm}
\begin{tenumerate}
\item In the text, identify the key goal underlying the Theme.
\item Is there is no apparent goal?   
\begin{itemize}
\item If yes, the opinion is probably non-utilitarian, so find and return a valence if any, but return no reason for it.  
\item If no, go to step 3. 
\end{itemize}
\item Determine whether the goal is satisfied:
\begin{itemize}
\item If yes, go to step 4,
\item If no, return a negative valence. 
\end{itemize}
\item Identify the subgoals involved in achieving the major goal.  
\item Identify how well the subgoals are satisfied.
\item Determine the final utilitarian sentiment based on the trade-off between different subgoals, and return it together with the trade-off analysis as the reasoning.  
\end{tenumerate}
\rule{7.5cm}{0.03cm}
This procedure requires the determination of the Goals or Subgoals and the Condition/Situation under which the opinion holder holds that opinion.  The former is discussed above; the latter can usually bet determined from the context of the given text.  

\subsection{Examples and Illustration}

As a running example we use simple restaurant reviews, sentences in italics indicating original text from the reviews\footnote{These reviews were originally from yelp reviews and revised by the authors for illustration purposes.}:

\paragraph{Case 1}
\begin{enumerate}
\item {\it My friends and I went to restaurant X}. 
\item {\it So many people were waiting there and we left without eating}.
\end{enumerate}
Following the algorithm sketch, the question {\bf ``was the major goal of going to a restaurant fulfilled?"} is answered {\bf no}.  
The reviewer is predicted to hold a negative sentiment.  Similar reasoning applies to Case~2 in Section~\ref{sec:intro}. 

\paragraph{Case 2}
\begin{enumerate}
\item {\it My friends and I went to restaurant X.} 
\item {\it The waiter was friendly and knowledgeable.} 
\item {\it We ordered curry chicken, potato chips and italian sausage.  The Italian sausage was delicious.}
\item {\it Overall the food was appetizing,} 
\item {\it but I just didn't enjoy the experience.} 
\end{enumerate}
To the question ``{\bf was the major goal of being full fulfilled?}" the answer is {\bf yes}, as the food was ordered and eaten.  Next the algorithms addresses the {\em how} (manner of achievement) question described in steps 4--6, which involves the functional elements of goals/needs embedded in each sentence:
\begin{enumerate}
\item {\it My friends and I went to restaurant X.} \\
Opinion Holder: I\\
Entity/Aspect/Theme: restaurant X\\
Need: sate hunger\\
Goal: visit restaurant \\
Sentiment Indicator: none\\
Valence: neutral
Condition: in restaurant X 
\item {\it The waiter was friendly and knowledgeable.} \\
Opinion Holder: I\\
Entity/Aspect/Theme: waiter\\
Need: gather respect/friendship \\
Subgoal: order food\\
Sentiment Indicator: friendly, knowledgeable\\
Valence: positive\\
Condition: in restaurant X
\item {\it We ordered curry chicken, potato chips and italian sausage. Italian sausage was delicious.}\\
Opinion Holder: I\\
Entity/Aspect/Theme: Italian sausage\\
Need: sate hunger\\
Subgoal: eat food\\
Sentiment Indicator: delicious\\
Valence: positive\\
Condition: in restaurant X
\item {\it Overall the food was appetizing,} \\
Opinion Holder: I\\
Entity/Aspect/Theme: food \\
Need: sate hunger\\
Subgoal: eat enough to remove hunger\\
Sentiment Indicator: appetizing\\
Valence: positive\\
Condition: in restaurant X
\item {\it but I just didn't enjoy the experience.} \\
Opinion Holder: I\\
Entity/Aspect/Theme: restaurant visit experience\\
Need: none --- this is not utilitarian\\
Goal: none\\
Sentiment Indicator: didn't enjoy\\
Sentiment Label: negative \\
Condition: in restaurant X
\end{enumerate}
The analysis of the needs/goals and their respective positive and negative valences allows one to justify the various sentiment statements, and (in the case of tie final negative decision) also indicate that it is not based on utilitarian considerations.   

\subsection{A Computational Model of Each Part}

Current computational models can be used to address each of the aspects involved in the sketch algorithm.  We provide only a high-level description of each. 
  
\paragraph{Deciding Functional Elements.}
Case 2 above involves three of the needs described in Maslow's hierarchy: food, respect/friendship, and emotion. 
The first two are stated to have been achieved.  
The third is a pure emotion, expressed without a reason, why the holder ``just didn't enjoy the experience". 
Pure emotions usually have no overt utilitarian value but only relate to the holder's high-level goal of being happy.  
In this example, we have to conclude that since all overt goals were met, either some unstated utilitarian Maslow-type need was not met, or the holder's opinion stems from a deeper psychological/emotional bias, of the kind mentioned in Section~\ref{sec:intro}, that goes beyond utilitarian value.  

\paragraph{Whether the Major Goal is Achieved.}
To make a decision about goal achievement, one must:
(1) identify the goal/subgoal of an action (e.g., buying the detergent, going to a restaurant); 
(2) identify whether that goal/subgoal is achieved.  
The two steps can be computed either separately or jointly using current machine learning models and techniques, including:
\begin{itemize}
\item {\bf Joint Model}: Annotate corpora for satisfaction or not for all goals and subgoals together, and train a single machine learning algorithm. 
\item {\bf Separate Model}: 
  \begin{enumerate}
  \item Determine the goal and its plans and subgoals either through annotation or as described in Section~\ref{sec:finding-goals}.
  \item Associate the actions or entities of the Theme (e.g., going to a restaurant; buying a car) with their respective (sub)goals. 
  \item Align each subgoal with indicator sentence(s) in the document (e.g., ``I got a small portion"; ``the car was all it was supposed to be"). 
  \item Decide whether the subgoal is satisfied based on indicator sentence(s).  
  \end{enumerate}
\end{itemize}

\paragraph{Learning Weights for Different Goals/Needs.}
One can clearly infer that the customer in case 2 assigns more weight to the emotional aspect, that being his or her final conclusion, 
and less to the food or respect/friendship (which comes last in this scenario).  
More formally, for a given text $D$, we discover $L$ needs/(sub)goals, with indices $1$, $2$,..., $L$. 
Each type of need/(sub)goal $i\in [1,L]$ is associated with a weight that contributes to the final sentiment valence decision $v_i$. 
In document $D$, each type of need $i$ is associated with achievement value $a_i$ that indicates how the need or goal is satisfied. 
The sentiment score $S_D$ for given document $D$ is then given by:
$$S_D=\sum_{i\in [1,L]} v_i\cdot a_i$$
This simple approach is comparable to a regression model that assigns weights to relevant aspects, where gold standard examples can be the overall ratings of the labeled restaurant reviews. 
One can view such a weight decision procedure as a supervised regression model by assigning a weight value to each discovered need.  
Such a procedure is similar to latent aspect rating introduced in \cite{wang2011latent,zhao2010jointly} by learning aspect weight (i.e., value, room, location, or service) for hotel review ratings. 
A simple illustrative example might be collaborative filtering in recommendation systems, e.g.,\cite{breese1998empirical,sarwar2001item}, 
optimizing need weight regarding each respective individual (which could be sampled from a uniform prior for humans' generally accepted weights). 

Since individual expectations can differ, it would be advantageous to maintain opinion holder profiles (for example, both yelp and Amazon keep individual profiles for each customer) that record one's long-term activity. 
This would support individual analysis of background, personality, or social identity, and enable learning of specific goal weights for different individuals.  

When these issues have been addressed, one can start asking deeper questions like: 
\begin{itemize}
\item {\it Q: Why does John like his current job though his salary is low?} \\
{\it A: He weighs employment more highly than family.}
\item {\it Q: How wealthy is a particular opinion holder?} \\
{\it A: He might be rich as he places little concern (weight) on money}. \\
\end{itemize}
or make user-oriented recommendations like:
\begin{itemize}
\item {\it Q: Should the system recommend an expensive--but-luxurious hotel or a cheap-but-poor hotel?} 
\end{itemize}

\subsection{Prior / Default Knowledge about Opinion Holders}

Sentiment/opinion analysis can be considerably assisted by the existence of a knowledge base that provides information about the typical preferences of the holder.

Individuals' goals vary across backgrounds, ages, nationalities, genders, etc.  An engineer would have different life goals from a businessman, or a doctor, a citizen living in South America would have different weighing systems from those in Europe or the United States, people in wartime  would have different life expectations from when in peacetime. 
Two general methods exist today for practically collecting such standardized knowledge to construct a relevant knowledge base: 
\begin{description}
\item [(1) Rule-based Approaches.]  Hierarchies of personality profiles have been proposed, and changes to them have long been explored in the social and developmental psychology literature, usually based on polls or surveys.  
For example, \shortcite{goebel1981age} found that children have higher physical needs than other age groups, love needs emerging in the transitional period from childhood to adulthood; esteem needs are the highest among adolescents; the highest self-actualization levels are found with adults; and the highest levels of security are found at older ages. 
As another example, researchers \cite{tang1998importance,tang2002effects,tang1997importance} have found that survival (i.e., physiological and safety) needs  dominate during wartime while psychological needs (i.e., love, self-esteem, and self-actualization) surface during peacetime, which is in line with our expectations.
For computational implementation, however, these sorts of studies provide very limited evidence, since only a few aspects are typically explored. 
\item[(2) Computational Inference Approaches.]
Despite the lack of information about individuals, reasonable preferences can be inferred from other resources such as online social media.  
A vast section of the Social Network Analysis research focuses on this problem, as well as much of the research of the large web search engine companies.  
Networking websites like Facebook, LinkedIn, and Google Plus provide rich repositories of personal information about individual attributes such as education, employment, nationality, religion, likes and dislikes, etc.  Additionally, online posts usually offer direct evidence for such attributes. 
Some examples include age \cite{rao2010classifying,rao2010detecting}, gender \cite{ciot2013gender}, living location \cite{sadilek2012finding}, and education \cite{mislove2010you}. 
\end{description}

\section{Conclusion and Discussion}

The past 15 years has witnessed significant performance improvements in training machine learning algorithms for the sentiment/opinion identification application.  
But little progress has been made toward a deeper understanding about what opinions or sentiments are, why people hold them, and why and how their facets are chosen and expressed. 
No-one can deny the unprecedented contributions of statistical learning algorithms in modern-day (post-1990s) NLP, for this application as for others.
However, ignoring cognitive and psychological perspectives in favor of engineering alone inevitably hampers progress once the algorithms asymptote to their optimal performance, since understanding {\em how} to do something doesn't necessarily lead to better insight about {\em what} needs to be done, or how it is best represented.  
For example, when inter-annotator agreement on sentiment labels peaks at 0.79 even for the rather crude 3-way sentiment granularity of positive/neutral/negative \cite{ogneva2010companies}, is that the theoretical best that could be achieved?  How could one ever know, without understanding what other aspects of sentiment/opinion are pertinent and investigating whether they could constrain the annotation task and help boost annotation agreement?  

In this paper, we described possible directions for deeper understanding, helping bridge the gap between psychology / cognitive science and computational approaches.  We focus on the opinion holder's underlying needs and their resultant goals, which, in a utilitarian model of sentiment, provides the basis for explaining the reason a sentiment valence is held.  (The complementary non-utilitarian, purely intuitive preference-based basis for some sentiment decisions is a topic requiring altogether different treatment.)  
While these thoughts are still immature, scattered, unstructured, and even imaginary, we believe that these perspectives might suggest fruitful avenues for various kinds of future work.  

\bibliographystyle{acl}
\bibliography{acl2013}  

\end{document}